\documentclass[conference]{IEEEtran}
\IEEEoverridecommandlockouts
\usepackage[ruled]{algorithm2e}
\usepackage{amsmath,amssymb,amsfonts}
\usepackage{algorithmic}
\usepackage{graphicx}
\usepackage{subfigure}
\usepackage{textcomp}
\usepackage{multirow}
\usepackage{hyperref}
\hypersetup{
	colorlinks=true,
	linkcolor=red,
	filecolor=blue,      
	urlcolor=green,
	citecolor=blue,
}

\usepackage{xcolor}

\def\BibTeX{{\rm B\kern-.05em{\sc i\kern-.025em b}\kern-.08em
    T\kern-.1667em\lower.7ex\hbox{E}\kern-.125emX}}
\begin{document}
\bibliographystyle{unsrt}
\title{Hardware Acceleration of Sampling Algorithms in Sample and Aggregate Graph Neural Networks\\
}







\author{\IEEEauthorblockN{Yuchen Gui}
\IEEEauthorblockA{\textit{School of Physical Sciences} \\
\textit{University of Science and Technology of China}\\
Hefei, China \\
guiyuchen@mail.ustc.edu.cn}
\and
\IEEEauthorblockN{Boyi Wei}
\IEEEauthorblockA{\textit{School of Physical Sciences} \\
\textit{University of Science and Technology of China}\\
Hefei, China \\
wby0823@mail.ustc.edu.cn}
\and
\IEEEauthorblockN{Wei Yuan}
\IEEEauthorblockA{\textit{School of Physical Sciences} \\
\textit{University of Science and Technology of China}\\
Hefei, China \\
yuanwei501240@mail.ustc.edu.cn}
\and
\IEEEauthorblockN{Xi Jin}
\IEEEauthorblockA{\textit{School of Physical Sciences} \\
\textit{University of Science and Technology of China}\\
Hefei, China \\
jinxi@ustc.edu.cn}
}

\maketitle

\begin{abstract}

Sampling is an important process in many GNN structures in order to train larger datasets with a smaller computational complexity. However, compared to other processes in GNN (such as aggregate, backward propagation), the sampling process still costs tremendous time, which limits the speed of training. To reduce the time of sampling, hardware acceleration is an ideal choice. However, state of the art GNN acceleration proposal did not specify how to accelerate the sampling process. What's more, directly accelerating traditional sampling algorithms will make the structure of the accelerator very complicated.  

In this work, we made two contributions: (1) Proposed a new neighbor sampler: CONCAT Sampler, which can be easily accelerated on hardware level while guaranteeing the test accuracy. (2) Designed a CONCAT-sampler-accelerator based on FPGA, with which the neighbor sampling process boosted to about 300-1000 times faster compared to the sampling process without it.
\end{abstract}


\section{Introduction}

Graph neural networks show excellent practicability in various situations, such as social networks, protein molecule structure prediction, citation networks and so on. Among various graph neural networks, some of them\mbox{\cite{bruna2014spectral}}\cite{duvenaud2015convolutional}\cite{defferrard2016convolutional}\cite{niepert2016learning}\cite{gcn} gather full graph information (such as Graph Laplacian) to learn parameters, which are limited by the fixed graph data structure and cannot be quickly applied to the ever-changing graph structure. Inductive representation neural networks\cite{graphsage}\cite{graphsaint} can solve this problem, which can be applied to rapidly changing data structures and shows high accuracy. At the same time, their various structures can be applied to different kinds of datasets, which have great practical value.


Another advantage of inductive representation neural networks is that they can gather graph local information very well. By aggregating local information from nodes' neighborhood, these networks avoid directly computing whole graph Laplacian, which can reduce computational complexity. In order to reduce computational complexity more, the sampling process was proposed. By sampling a given number of neighbor nodes and only aggregating their information while training, sample-aggregate networks can make it possible to train very large graph at a relatively low cost while guaranteeing test accuracy. However, compared to other processes of GNN, sampling process still consumes much time and computing resources, especially on large datasets. For example, if we train Reddit dataset with traditional GraphSAGE network
\footnote{Tested on Intel(R) Xeon(R) Platinum 8260 CPU @ 2.40GHz with NVIDIA Tesla P100. The test code is built with PyTorch Geometric (PyG)\cite{PyG}, using traditional stochastic neighbor sampler (NeighborLoader) with num\_neighbors=[25, 10], batch\_size=1024.}
, we will find that the sampling process takes more than 100 times longer than other GNN processes like aggregate, update, and so on. This is because the traditional sampling process requires frequent access to the dataset and its edge index. Therefore, hardware acceleration for the sampling process is necessary.


Currently, there are several hardware acceleration proposals for GNN. For example, HyGCN\cite{hygcn} proposed a hardware accelerator using an aggregation engine and a combination engine to exploit various parallelism and reuse highly reusable data efficiency. At the same time, AWB-GCN\cite{awbgcn} was proposed to monitor the sparse graph pattern and adjust the workload distribution among a large number of processing elements; GCNAX\cite{gcnax} optimized dataflow for GCNs that simultaneously improves resource utilization and reduces data movement; However, except HyGCN, these accelerators did not contain a module that accelerates sampling process. 
Even though HyGCN contains a sampling module, it did not show how it works in a detailed way and the performance of the sampling accelerator. Therefore, we decided to design a novel sampling accelerator, it should be simple, yet performs well.

There are several state-of-the-art sampling algorithms that can guarantee test accuracy well, like GraphSAGE traditional stochastic sampler\cite{graphsage}, GraphSAINT sampler\cite{graphsaint}, adaptive sampler\cite{adaptivesampler} and so on. However, if we want to accelerate these processes with hardware, the structure of hardware may be very complicated (see section \ref{hardware}). Therefore, a new sampling algorithm is needed in order to reduce the complexity of the hardware accelerator.

In this work, We proposed a novel sampling algorithm, CONCAT (CONCATenate) sampler, which creatively concatenates low-level-sample-graph into higher-depth-sample-graph, making it easier to be accelerated with hardware (FPGA) while ensuring the test accuracy. We also designed a hardware accelerator for CONCAT sampler, with which the sampling process can be boosted to about 300-1000 times (see section\ref{hardwaretest}) compared to the traditional only-software process.

\begin{figure*}[htbp]
    \centering
    \includegraphics[width=16cm]{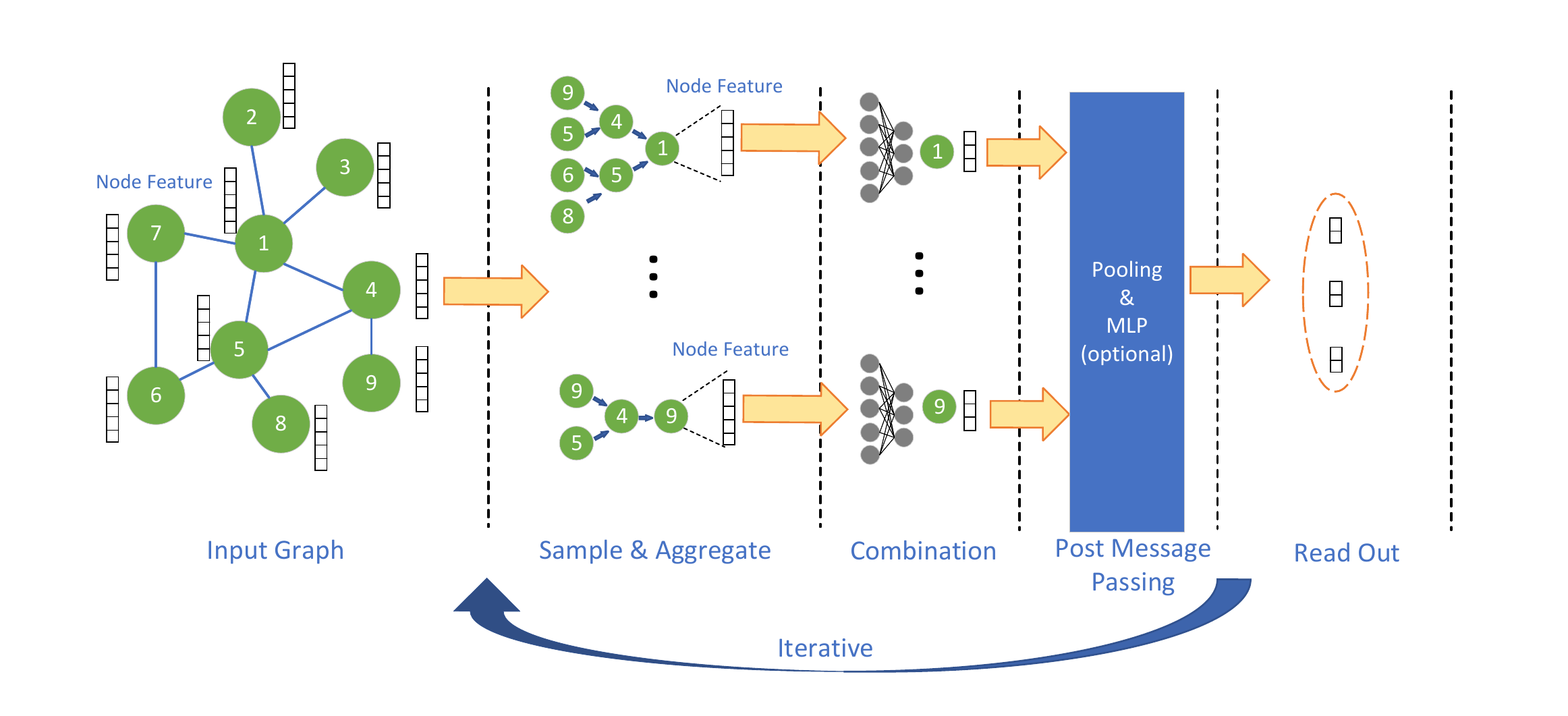}
    \caption{Schematic structure of GCN/GraphSAGE model}
    \label{GCN}
\end{figure*}

\section{Background and Related work}
\subsection{Notations and Definitions}
Here we recap some basic concepts of Graph. Notations and their explanations are listed below. We also want to clarify the difference between $K$-neighborhood and $K$-hop neighborhood. In this paper, we refer $K$-neighborhood as a set of neighbor nodes $v_j$ with dist($v_i, v_j)\leqslant K$, and refer $K$-hop neighborhood as a set of neighbor nodes $v_j$ with dist($v_i, v_j)=K$.
\begin{table}[htbp]
\caption{Notations and definition about GNN}
\resizebox{\linewidth}{!}{
\begin{tabular}{cc}
\hline
Notation               & definition                                                                                               \\ \hline
$\mathcal{G}$          & (un)directed (attributed) graph                                                                           \\
$\mathcal{V}$          & set of nodes in graph $\mathcal{G}$                                                                       \\
$\mathcal{E}$          & set of edges in graph $\mathcal{G}$                                                                       \\
$A$                    & adjacency matrix of graph $\mathcal{G}$                                                                    \\
$L$                    & Laplacian matrix of graph $\mathcal{G}$                                                                   \\
$v_i$                  & nodes of graph $v_i\in\mathcal{V}$                                                                        \\
$h_{v_i}$               & feature vector of node $v_i$                                                                              \\
$d_{v_i}$               & degree of node $v_i$                                                                                      \\
$X$                     & feature matrix of graph (composed by feature vectors)                                                     \\
$\mathrm{dist}(v_i,v_j)$ & distance between node $v_i$ and node $v_j$                                                              \\
$\mathcal{N}_K(v_i)$   & $K$-neighborhood $\mathcal{N}_K(v_i)=\{v_j\in \mathcal{V}|\text{dist}(v_i, v_j)\leqslant K\}$              \\
$\mathcal{S}_K(v_i)$   & Sampling subset of $\mathcal{N}_K(v_i)$\\
$\mathcal{N}_K'(v_i)$   & $K$-hop neighborhood $\mathcal{N}_K'(v_i)=\{v_j\in \mathcal{V}|\text{dist}(v_i, v_j)= K\}$                 \\      
$\mathcal{S}_K'(v_i)$   & Sampling subset of $\mathcal{N}_K(v_i)$\\
$\mathcal{G}_{v_i, K}$ & $K$- computational graph of node $v_i$ with$v_j\in\mathcal{N}_K(v_i)$                                                                    \\ \hline
\end{tabular}
}

\end{table}

\begin{figure}[htbp]
\centering
\centerline{\includegraphics[width=8cm]{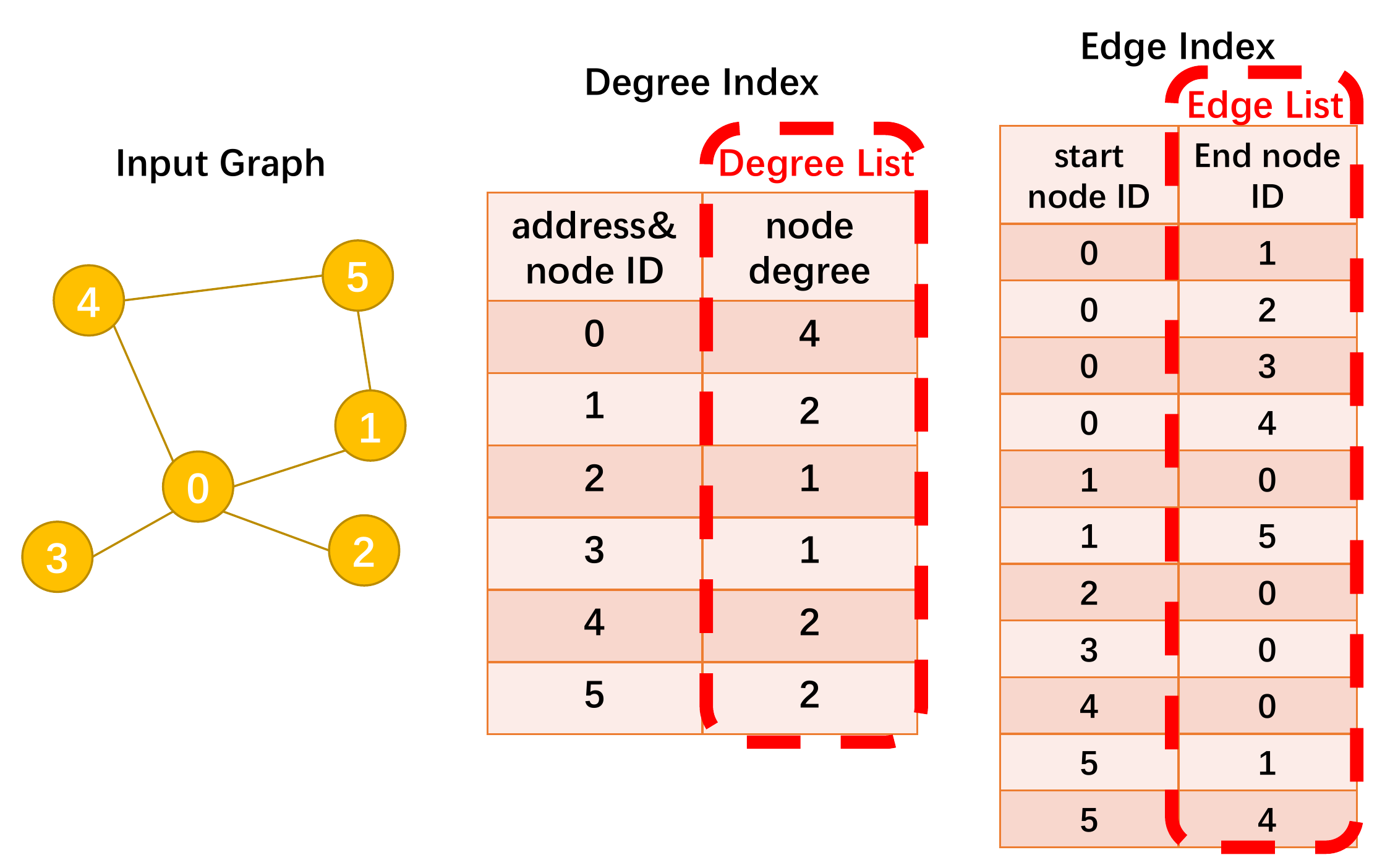}}
    \caption{The relationship between degree index and degree list, and the relationship between edge index and edge list. We only store degree list and edge list in RAM.}
    \label{degree_edge_list}
\end{figure}

For sampled nodes from the dataset, their basic information is required, including node degree and neighbor nodes. We use a column of serial data to represent each node's degree, as shown in  Fig. \ref{degree_edge_list}, we call it "degree list". They are stored in RAM, using node IDs as their corresponding offset address. 

Meanwhile, we use "edge index" to represent neighborhood information from each node, in which the first column represents start node ID and the second column represents end node ID. They are sorted according to start node's ID number. As a matter of fact, in sampling process, we only need to store second column, as they are sorted in order and we can locate them via node ID and degree information. Therefore, we only store the second column in RAM and called them "edge list", as shown in Fig. \ref{degree_edge_list}. 

\subsection{GNN models}
Here we will introduce the basic information of some GNN models. 
\subsubsection{\textbf{Graph Convolutional Network (GCN)}}

Graph Convolutional Network is one of the most classical graph neural networks which successfully applies convolutional network on graph learning.\cite{gcn}\cite{wu2020comprehensive}. Its inference model can be described as:
\begin{equation}
    \left\{\begin{array}{l}
a_{v}^{k}=\displaystyle{\sum_{u \in \mathcal{N}_1(v) \cup\{v\}} \frac{h_{u}^{k}}{\sqrt{d_{u} \cdot d_{v}}}} \\
h_{v}^{(k+1)}=\operatorname{ReLU}\left(W^{k} a_{v}^{k}+b^{k}\right)
\end{array}\right.
\end{equation}

\subsubsection{\textbf{GraphSAGE}}
Based on GCN, GraphSAGE adopted a sampling process in training in order to alleviate the exponential growth of computational complexity on large datasets when using GCN\cite{graphsage}. In GraphSAGE, the forward propagation process can be described as:
\begin{equation}
    \left\{\begin{aligned}
a_{v}^{k} &=\sum_{u \in \mathcal S_1(v) \cup\{v\}} \frac{h_{u}^{k}}n \\
h_{v}^{(k+1)} & = \operatorname{ReLU}\left(W^{k} a_{v}^{k}+b^{k}\right)
\end{aligned}\right.
\end{equation}
Where n is the number of nodes sampled from $v_i$'s neighborhood. In this work, we mainly focus on GraphSAGE model. Our proposed sampling algorithm and hardware acceleration are applied and tested based on GraphSAGE model.

In fact, GCN and GraphSAGE can be described in a uniform equation\cite{gcnax}:
\begin{equation}
    X^{(k+1)}=\sigma\left(\hat{A} X^{(k)} W^{(k)}\right)
\end{equation}
For GCN, $\hat{A}=D^{-\frac{1}{2}}(A+I)D^{\frac12}$. For GraphSAGE, $\hat{A}$ denotes adjacency matrix of sampled nodes with a scaling factor of $1/n$.

\begin{figure*}[htbp]
    \centering
    \includegraphics[width=14cm]{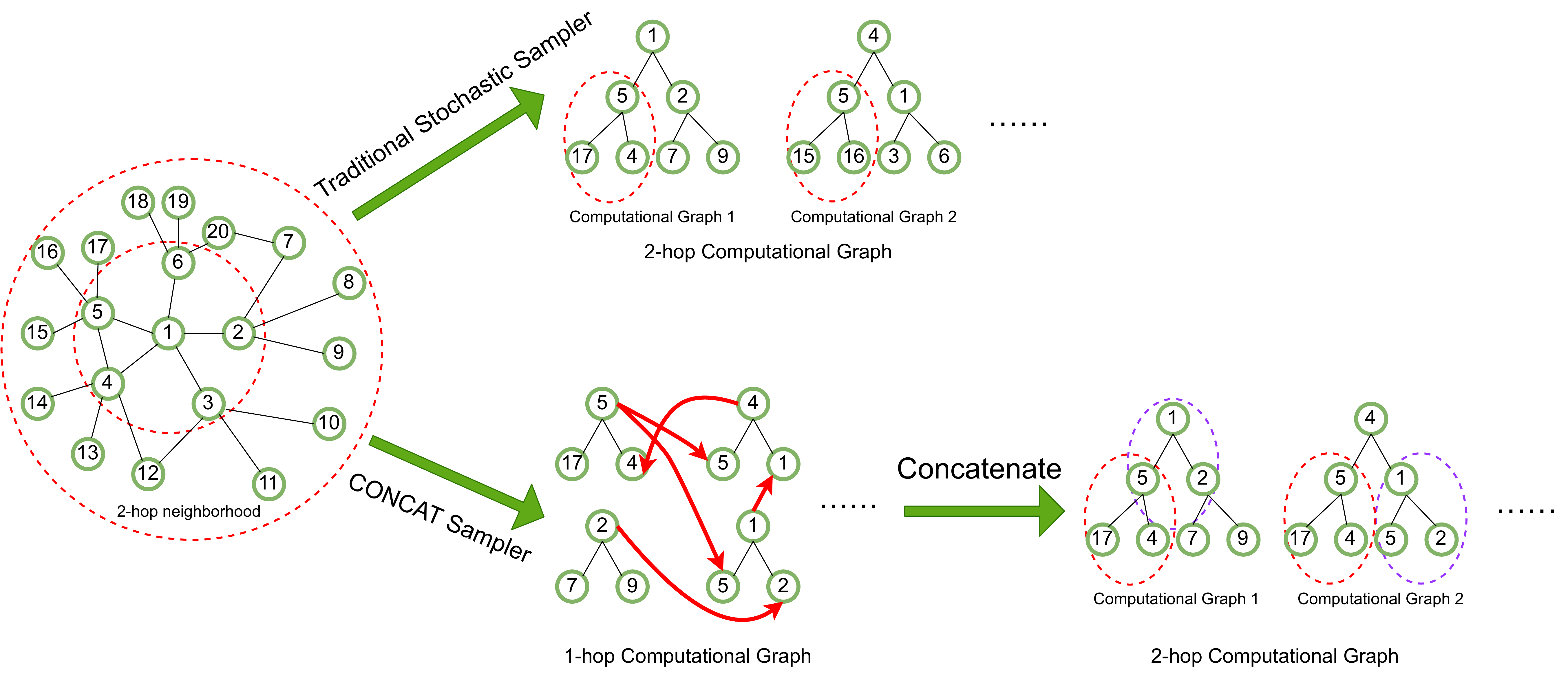}
    \caption{Comparison between traditional stochastic neighbor sampler and our CONCAT sampler. From the figure, we can see that in computational graph sampled by the traditional stochastic neighbor sampler, each neighborhood is sampled independently, while CONCAT sampler only samples and generate 1-neighborhood computational graphs and concatenates within them to generate 2-or-higher-neighborhood computational graphs.}
    \label{sample}
\end{figure*}

\subsection{Neighbor Sampling methods}
Before our work, there are several neighbor sampling methods proposed. Here we want to introduce two of them that have relationship with our work
\subsubsection{\textbf{Stochastic (Random) Sampler}}
Traditional Stochastic (Random) Sampler\cite{graphsage} samples neighbor nodes with uniform distribution, which means the possibility that a node being sampled is $1/n$, where $n$ is the number of nodes in the neighborhood, like Fig.\ref{sample} shown.
\subsubsection{\textbf{Importance Sampler}}
Importance Sampler is firstly proposed in FastGCN\cite{chen2018fastgcn}. In Importance Sampler, it assumes that every node has the same distribution $q$, and samples neighbor nodes according to $q$. In fact, the Stochastic Sampler is a special case of Importance Sampler. In Stochastic Sampler, every node has uniform distribution, which means $q\thicksim U[0, 1]$.

 All of these samplers presented impressive performance and reduced the cost of training GNNs, which makes it possible to train very large graphs. However, from our perspective, if we want to accelerate the sampling process via FPGA, more considerations are needed. 
 For example, if we sample nodes in n-neighborhood ($n>1$), the sampled edge list will be stored in RAM, and the sampler will sample in n-hop neighborhood according to the sample result in (n-1)-hop-neighborhood. The constant access to edge list (which means constant access to RAM) will cause the hardware structure very complicated, and the limited space of on-chip memory will restrict the performance of the accelerator on large datasets. Therefore, we need to design a new sampler, which can cater to small on-chip memory space even in large datasets. At the same time, the test accuracy shouldn't be reduced a lot.

\section{Proposed a new sampler: CONCAT Sampler}
In order to solve the problems mentioned above, we proposed a novel yet effective sampler---CONCATenate (CONCAT) sampler. In CONCAT sampler, neighbors are sampled following the way below: 
\begin{enumerate}
    \item Sample in 1-neighborhood to generate the original computational graphs $\mathcal{G}_{v_i, 1}$;
    \item Concatenate the sample result into the original computational graphs;
    \item Generate 2-neighborhood computational graphs $\mathcal{G}_{v_i, 2}$;
    \item Iterate the sample methods above to get a k-neighborhood computational graphs $\mathcal{G}_{v_i, K}$.
\end{enumerate}
\begin{algorithm}
\caption{CONCAT Sampler sampling algorithm}
\LinesNumbered 
\KwIn{Graph $\mathcal{G}(\mathcal{V}, \mathcal{E})$; Node $v\in\mathcal{V}$; Search depth $K$; number of neighbors $N$}
\KwOut{$K$-neighborhood computational graph for node $v$: $\mathcal{G}_{v, K}$}
\For{$i=1...N$}{
    RandomSample($u_i\in\mathcal{N}_1(v)$)\;
}
$\mathcal{G}_{v,1}\leftarrow \mathcal{G}(v, u;(v, u))$\;
\For{$j=2...K$}{
    $\mathcal{G}_{v,j}\leftarrow \mathrm{CONCAT}(\mathcal{G}_{v, j-1}, \mathcal{G}_{v, 1})$\;
}
\end{algorithm}

To be more specific, we can compare CONCAT sampler with the traditional sampler, like GraphSAGE stochastic sampler, as Fig. \ref{sample} shown. If we sample 2 neighbor nodes for each node and search in depth $K=2$, we can generate computational graphs for each node. We will find in each computational graph, every node is sampled at random independently. For example, in node 1 and 4's computational graph, node 5 are sampled, however, the neighbor nodes that are sampled from node 5 are totally different (node 17, 4, and node 15, 16). 
Different from the traditional samplers, CONCAT sampler samples neighbor nodes in a totally new way. In fact, it only samples in 1-neighborhood, concatenate within $\mathcal{G}_{v, 1}$ to generate higher-neighborhood computational graph. Like Fig.\ref{sample} shown, the node 1 and 4's $\mathcal{G}_{v, 2}$ are derived from the concatenation of node 1, 2, 4, and 5's $\mathcal{G}_{v, 1}$. From the experiment result mentioned later, CONCAT sampler can guarantee the test accuracy while simplifying the process of acceleration with FPGA.

\section{Hardware Acceleration of sampling process}\label{hardware}

So far, few works mainly focus on hardware acceleration of neighbor sampling process, most of them concern about the whole acceleration of training pipeline and regard neighbor sampling as a step of data preprocessing, or implement sampler in the architecture without giving more detailed description\cite{zhang2021efficient}\cite{liu2021gnnsampler}\cite{abadal2021computing}. Some other works \cite{liu2020bandit}\cite{zhang2021biased}\cite{zhou2022tgl}\cite{hao2021pre}\cite{yan2021progressive} proposed some new samplers or sampling methods from software level without thinking about acceleration from hardware level. In this work we proposed a hardware architecture in order to accelerate our CONCAT sampler and use FPGA as out platform.

As mentioned above, in sampling process, the information we need includes target nodes' degree and node IDs that have connection with target nodes, and they are stored in "degree list" and "edge list" in RAM, respectively. In degree list, the nodes' degree are stored according to node IDs, in other words, we use node ID as offset address. In edge list, neighbor node IDs are also stored according to target node ID (For undirected graph, we use two directed edges opposite in direction to represent one undirected edge). Therefore, if we use $x_d$ to represent the address that stores node $i$'s degree, use $x_e$ to represent the address that stores node $i$'s 1-hop neighborhood's node ID, and $x_{d0}$, $x_{e0}$ represent degree list's base address and edge list's base address, $x_d$ and $x_e$ can be represented as:
\begin{equation}
    x_d=x_{d0} + d_{vi}
\end{equation}
\begin{equation}
    x_{e0} + \sum_{n=0}^i d_{vn} \leqslant x_e<x_{e0} + \sum_{n=0}^{i+1} d_{vn}
\end{equation}

\begin{figure*}[htbp]
    \centerline{\includegraphics[width=16cm]{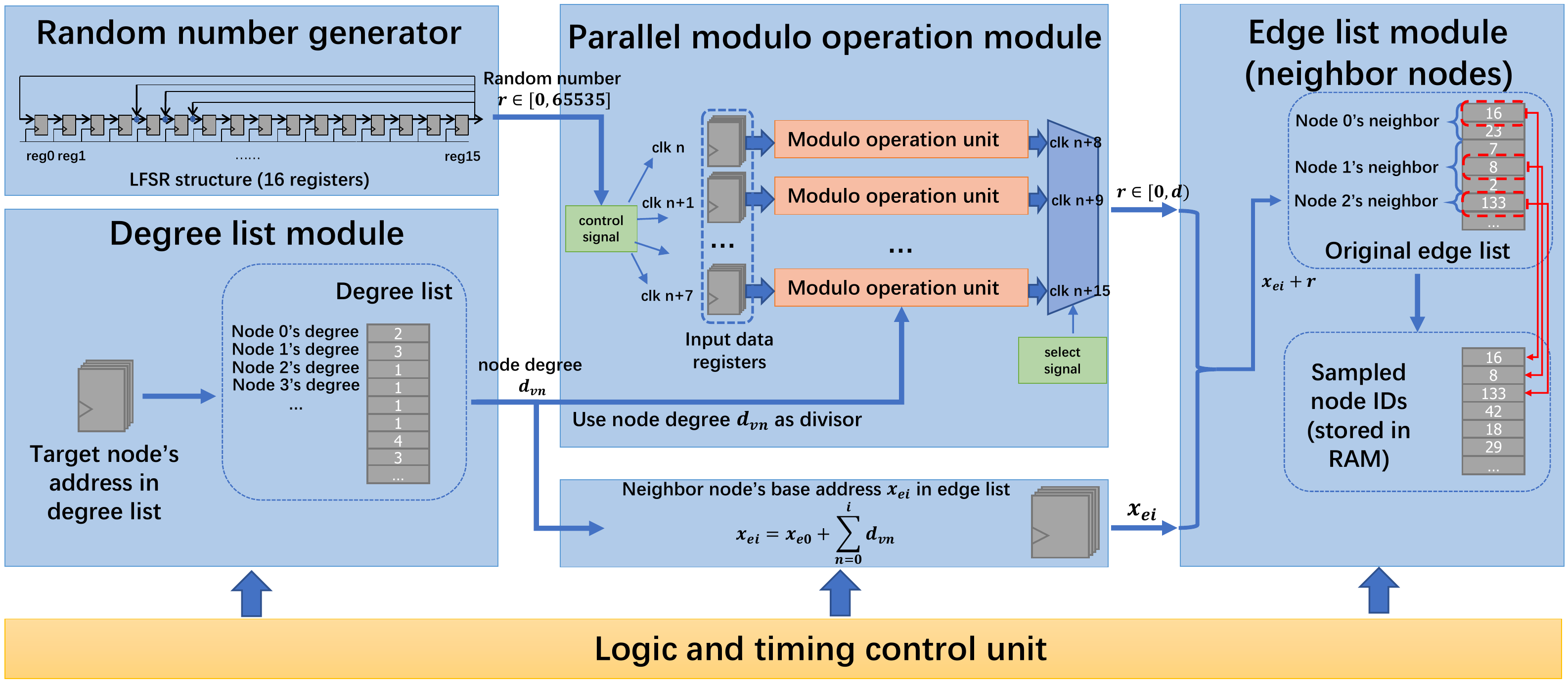}}
    \caption{Structure of the small datasets sampling accelerator: The degree list of the graph is stored in the degree data module in RAM. The edge list contains the each node's 1-hop neighbor node IDs. In the sampling process, the sampler will randomly sample given number of neighbor node, and output sampled node IDs.}
    \label{small_acce}
\end{figure*}

\begin{figure}[htbp]
    \centering
    \includegraphics[width=8cm]{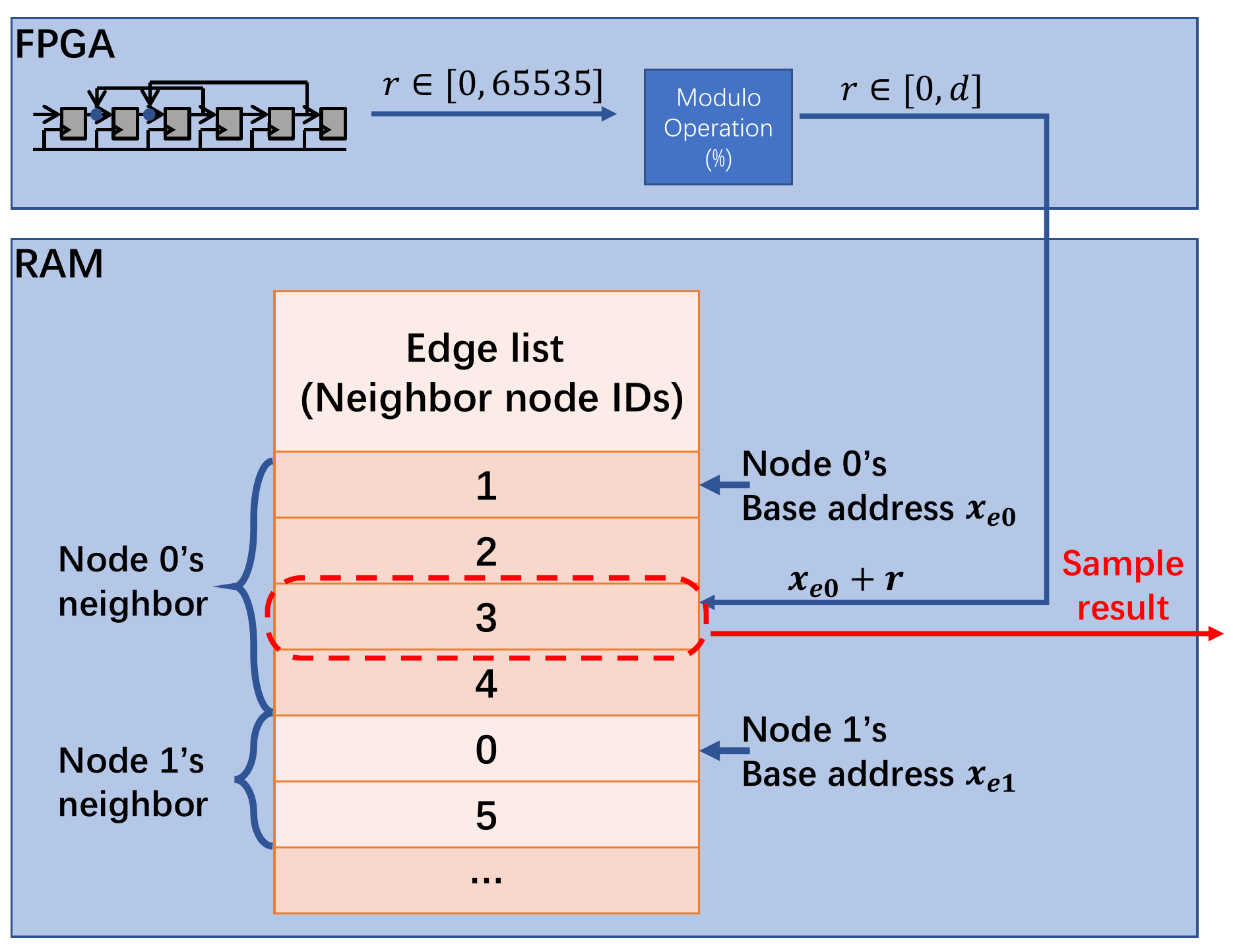}
    \caption{A schematic diagram of sampling process. Random number generated from FPGA works as offset address. Address that stores sampled node ID equals the sum of base address and offset address.}
    \label{sample_shematic}
\end{figure}
Due to the limited BRAM size in FPGA, hardware structure should change according to the size of input dataset. For the small datasets like Cora, Citeseer, PubMed and so on, the on-chip BRAM can entirely store their edge lists and degree lists, which enables sampler directly read full information from it, and complete sampling process for a single time. However, for larger datasets like NELL, ogbn-arxiv\cite{hu2020ogb}, Reddit and so on, the on-chip memory cannot store whole edge lists and degree lists, due to which their degree lists and edge lists must be stored in off-chip memory like DDR, and they can be load into FPGA in sampling process. In sampling process of large datasets, for each time the sampler loads a batch of data lists and edge lists, samples a given number of neighbor nodes in and outputs the sample result. The data lists and edge lists are loaded sequentially, according to node IDs.

\begin{figure*}[htbp]
    \centerline{\includegraphics[width=16cm]{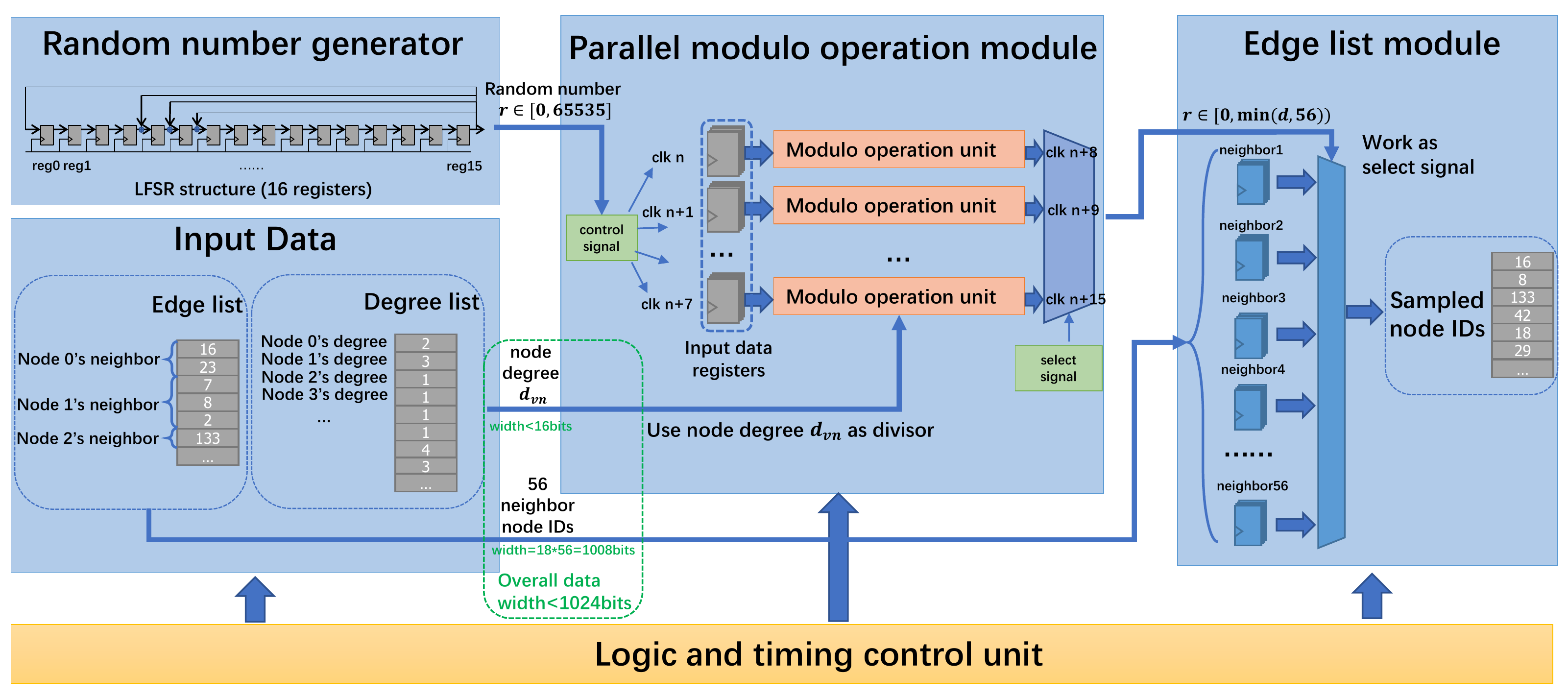}}
    \caption{Structure of sampling module for large datasets (use Reddit as an example). 56 neighbor node IDs are stored in 56 register arrays, with node's degree information stored in an extra register array. Random number selects a particular register array, and the node ID stored in this register array is the sample result in this clock cycle.}
    \label{large_acce}
\end{figure*}

\subsection{Sampling module for small datasets (Fig.\ref{small_acce})}
For small datasets (Cora, Citeseer, Pubmed etc.), the sampling process mainly includes: 

(1) Generate random number $r$.

(2) $r$ modulo $d_{vi}$($r = r\mod d_{vi}$) to make sure the random number does not exceed the degree of the node.

(3) The address of node ID of neighbor node going to be sampled can be computed according to $\displaystyle{x_{e0}+\sum_{n=0}^i d_{vn} + r}$, and the sampler will read from target address and output the result to the specific port.


As a kind of stochastic sampler, CONCAT sampler needs a random number generator. However, for hardware devices, a completely-true random number generator is actually difficult to be implemented. Therefore, we designed a pseudo-random number generator. Many types of random number generators use linear congruence algorithm, their periods are very long in order to make the generated numbers seem completely random. For linear congruence algorithm, it consists a lot of modulo operation, which is complicated and slow in hardware devices. If we implement this method in FPGA, it will consume a large amount of resources and will let FPGA work at a lower frequency. To solve this problem, we choose LFSR (Linear-feedback shift register) structure to generate random number. In this strucure, we used 16 registers so the original generated random number is between 0 and 65535. It is important to note that we still used modulo operation after the random number is generated. We only avoid modulo operation in the generation process.


After the generation of random number, we used modulo operation to make the range of the random number is between 0 and $d_{vi}$, random number bigger than node's degree is meaningless in sampling process. As mentioned above, modulo operation is a long time process in FPGA. To solve this problem, we used 8 parallel modulo operation modules in our architecture. They work in parallel, and take turns to input and output data to improve the frequency. The process is controlled by the logic and timing control unit. Based on this way, we can generate a random number between 0 and $d_{vi}$ in one clock cycle.

After modulo operation, as Fig.\ref{sample_shematic} shown, if we note $x_{ei}=x_{e0}+\displaystyle{\sum_{n=0}^i d_{vn}}$ as the base address of node $v_i$, and random number $r$ works as offset address, then the address that stores neighbor node ID being sampled equals $x_{ei} + r$. After sampled the given number of neighbor nodes of node $v_i$, the base address changed to $x_{e(i+1)}$ and the sampler will start the sampling process of node $v_{i+1}$. In the end, the sampled result can be sequentially sent to a given port or stored in RAM, depends on user's intention.



\subsection{Sampling module for large datasets (Fig.\ref{large_acce})}

    

Sampling of large datasets requires data from off-chip sources (See Chapter \ref{hardware}). Off-chip data can be read from IO ports and stored into data registers, and sampling module samples the data stored in registers. Here we use Reddit dataset as an example to introduce our solution of sampling from large datasets.
Due to the limited bus width, the maximum bit width of data transmitted into FPGA at one time is 1024 bits. However, for Reddit dataset, it has a large number of nodes (232965 nodes), which means that we need 18 bits number to store node IDs. Therefore, for a single time, the sampler can only load 56 neighbor node IDs, with 16 bits width free regarding the maximum 1024 bits. We use the free bus width to transmit node degree. As a matter of fact, some nodes have more than 56 neighbor nodes, in other words, we cannot load all neighbor nodes at once. 
To solve this problem, we found that we can simply discard the neighbors other than the first 56 neighbors when the number of neighbors exceeds the limit and then sample among only 56 neighbors. Due to the reduction in the number of neighbors that should participate in sampling, it is necessary to verify whether this solution will lead to an unacceptable decrease in accuracy before using it. The data in Fig.\ref{acc_comp}(a) shows that the test accuracy of this solution does not decrease, which indicates that this solution is feasible.

Different from the sampler of small datasets, 56 neighbor node IDs are stored in 56 register arrays. Random number is generated with the range of 0 and $\min(d_{vi}, 56)$, and it selects the corresponding neighbor node ID as sample result. After sampling given number of neighbor nodes, the sampler will read the next node's 56 neighbor IDs into register, and repeat the process mentioned above.

\section{Experiments}

\subsection{Accuracy of CONCAT Sampler and its comparison with the traditional 2-neighborhood sampler}
\begin{table}[htbp]
\caption{Hyperparameter Settings}
\resizebox{\linewidth}{!}{
\begin{tabular}{|c|c|c|c|c|}
\hline
Dataset   & Optimizer             & aggregation           & learning rate & Hidden layer dim \\ \hline
Cora      & \multirow{5}{*}{Adam} & \multirow{5}{*}{mean} & 0.001         & 32               \\ \cline{1-1} \cline{4-5} 
Citeseer  &                       &                       & 0.01          & 32               \\ \cline{1-1} \cline{4-5} 
Pubmed    &                       &                       & 0.01          & 32               \\ \cline{1-1} \cline{4-5} 
NELL    &                       &                       & 0.001          & 512               \\ \cline{1-1} \cline{4-5} 

OGB-arxiv &                       &                       & 0.001         & 256              \\ \cline{1-1} \cline{4-5} 
Reddit    &                       &                       & 0.001         & 64               \\ \hline
\end{tabular}}
\label{hyperparameter}
\end{table}

To fully show the characteristics of the training results of CONCAT Sampler, we selected six datasets with different size to test the model's performance. Among these datasets, Cora, Citeseer, and PubMed have a relatively small number of nodes and edges while NELL, OGB-arxiv and Reddit are large scale datasets. (see table \ref{acc_table})

In our test, we mainly focus on the accuracy difference between CONCAT sampler and traditional 2-neighborhood stochastic sampler, for our sampler is proposed with the goal of easy to be accelerated while guarantee test accuracy. Therefore, our sampler is not designed to improve test accuracy. Compared to traditional stochastic sampler, it only need to ensure the accuracy fluctuates in an acceptable range. (Accuracy loss less than 1.5\%).

To apply CONCAT sampler into training and evaluation process, we designed a dataloader derived from \verb|torch_geometric.loader.NeighborSampler| (A class in Pytorch Geometric \cite{PyG} to generate mini-batch in training process). It uses CONCAT sampler to sample from each node, and generate mini-batch in train, test and validation process. Our hyperparameter settings are listed in table \ref{hyperparameter}.



\begin{figure}
    \centering
    \includegraphics[width=8.5cm]{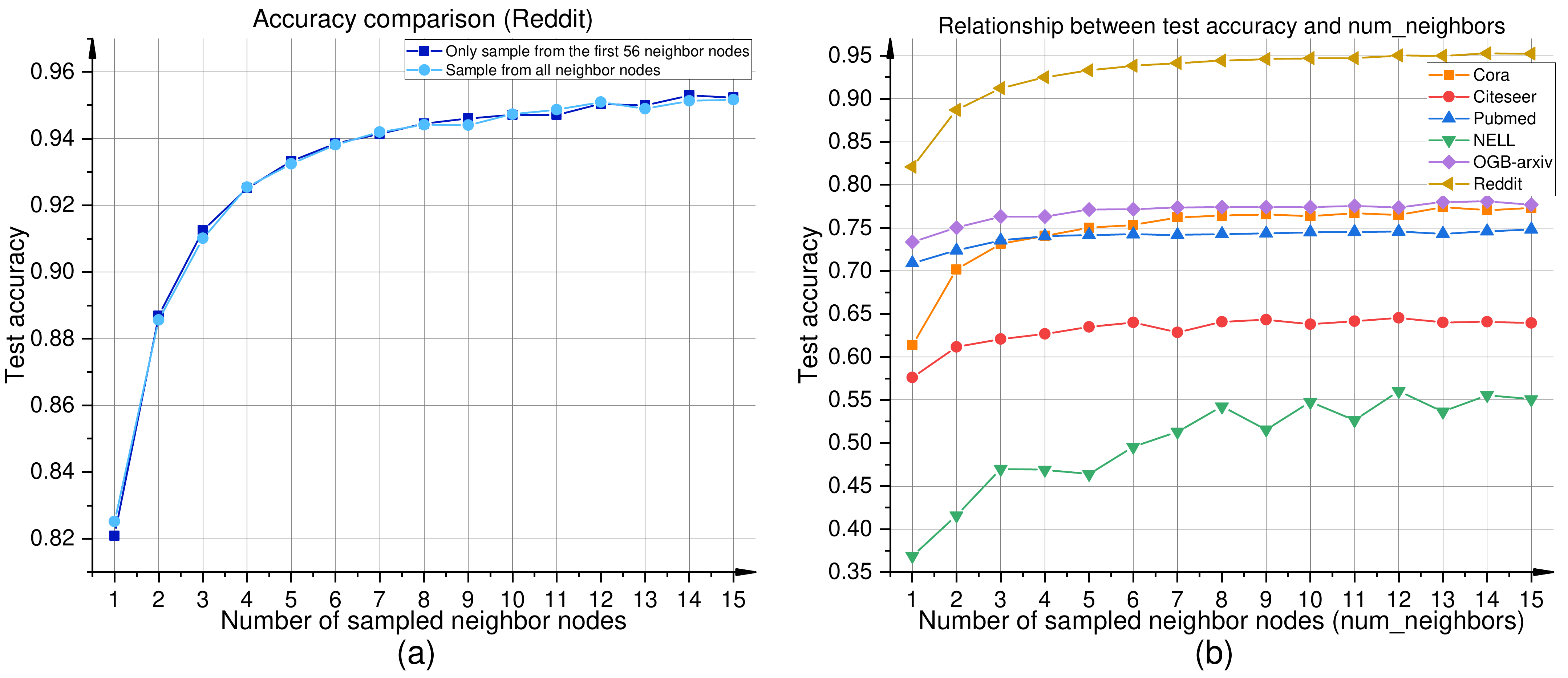}
    \caption{(a): Test accuracy comparison between two different sample strategy (Only sample from the first 56 neighbor nodes and sample from all neighbor nodes, take Reddit as an example). (b): Relationship between test accuracy and number of sampled neighbor nodes (use CONCAT sampler).}
    \label{acc_comp}
\end{figure}

Obviously, for each dataset, their test accuracy will be higher if more neighbor nodes are sampled in the training process. In this work, we tested the relationship between test accuracy and number of sampled neighbor nodes (\verb|num_neighbors|, from 1 to 15). The test result are shown in Fig.\ref{acc_comp}(b). From the result we can find the accuracy will converge when \verb|num_neighbors| reaches to about 15. In table \ref{acc_table} we listed different datasets' test accuracy when \verb|num_neighbors=15|, and compared test accuracy between CONCAT sampler and 2-neighborhood sampler. From the result we can find that  the accuracy loss of the CONCAT sampler is less than 1.5\%, thus we can say that CONCAT sampler can keep the accuracy within acceptable limits. Specifically, for Reddit, if we sample [12,12] neighbors with CONCAT sampler, we can reach the same accuracy (0.950) as if we had sampled [25,10] neighbors with 2-neighborhood sampler.


\begin{table}[]
\caption{Accuracy on Different Datasets}
\resizebox{\linewidth}{!}{
\begin{tabular}{|c|cccc|}
\hline
\multirow{2}{*}{\textbf{\begin{tabular}[c]{@{}c@{}} \\ Dataset\end{tabular}}} & \multicolumn{4}{c|}{\textbf{Datasets Information and Test Results}}                                                                                                                                                                                                                                               \\ \cline{2-5} 
                                                                                   & \multicolumn{1}{c|}{\textit{\textbf{Nodes}}} & \multicolumn{1}{c|}{\textit{\textbf{Edges}}} & \multicolumn{1}{c|}{\textit{\textbf{\begin{tabular}[c]{@{}c@{}}Accuracy with\\ 2-neighborhood sampler\end{tabular}}}} & \textit{\textbf{\begin{tabular}[c]{@{}c@{}}Accuracy with \\ CONCAT sampler\end{tabular}}} \\ \hline
Cora                                                                               & \multicolumn{1}{c|}{2708}                    & \multicolumn{1}{c|}{5429}                    & \multicolumn{1}{c|}{0.773}                                                                                     & \textbf{0.773}                                                                            \\ \hline

Citeseer                                                                           & \multicolumn{1}{c|}{3327}                    & \multicolumn{1}{c|}{4732}                    & \multicolumn{1}{c|}{0.639}                                                                                     & \textbf{0.640}                                                                            \\ \hline

PubMed                                                                             & \multicolumn{1}{c|}{19717}                   & \multicolumn{1}{c|}{44338}                   & \multicolumn{1}{c|}{0.747}                                                                                     & 0.746                                                                                     \\ \hline

NELL                                                                             & \multicolumn{1}{c|}{65755}                   & \multicolumn{1}{c|}{251550}                   & \multicolumn{1}{c|}{0.5641}                                                                                     & 0.5507                                                                                     \\ \hline

OGB-arxiv                                                                             & \multicolumn{1}{c|}{169343}                   & \multicolumn{1}{c|}{ 1166243}                   & \multicolumn{1}{c|}{0.7796}                                                                                     & 0.7769                                                                                    \\ \hline

Reddit                                                                             & \multicolumn{1}{c|}{232965}                   & \multicolumn{1}{c|}{ 57307946}                   & \multicolumn{1}{c|}{0.9547}                                                                                     & 0.9528                                                                                    \\ \hline

\end{tabular}
}
\label{acc_table}
\end{table}



\subsection{Hardware acceleration test}\label{hardwaretest}

We used Xilinx Virtex®-7 FPGA VC707 Evaluation Kit as our hardware platform. We applied our proposed acceleration architecture on this device, and conducted several experiments to test the performance of hardware accelerator. In our experiments, the sampling accelerator works at a maximum frequency of 250MHz.


The module can sample one neighbor node in each clock cycle. The number of sampled neighbor nodes can be adjusted based on model configuration. In our experiment, \verb|num_neighbors| is set to 15 according to test result mentioned above, and other hyperparameter settings are the same as table \ref{hyperparameter}. Under this samples number setting, the acceleration test results are listed in the table \ref{acceleration_data}.

From the results we can find the sampling process has been greatly accelerated (about 300 to 1000 times faster). It also proves that our proposed CONCAT sampler is easy to be accelerated on hardware device.


\begin{table}[]
\caption{Acceleration Results}
\resizebox{\linewidth}{!}{
\begin{tabular}{|c|c|c|c|}
\hline
\textbf{Dataset} & \textbf{\begin{tabular}[c]{@{}c@{}}Sampling Time \\ (Software) \\ (ms)\end{tabular}} & \textbf{\begin{tabular}[c]{@{}c@{}}Sampling Time \\ (Hardware) \\ (ms)\end{tabular}} & \textbf{Speed Comparison} \\ \hline
Cora             & 65.38                                                                                 & 0.162                                                                                 & $\times$ 404              \\ \hline

Citeseer         & 81.11                                                                                 & 0.200                                                                                 & $\times$ 406              \\ \hline

PubMed           & 467.31                                                                                & 1.183                                                                                 & $\times$ 395              \\ \hline

NELL           & 1254.97                                                                                & 3.945                                                                                 & $\times$ 318              \\ \hline

OGB-arxiv           &    3441.9                                                                              &   10.161                                                                              & $\times$ 339            \\ \hline

Reddit           & 14076                                                                                 & 13.978                                                                                & $\times$ 1007             \\ \hline


\end{tabular}
}
\label{acceleration_data}
\end{table}

\begin{figure}[htbp]
    \centerline{\includegraphics[width=8cm]{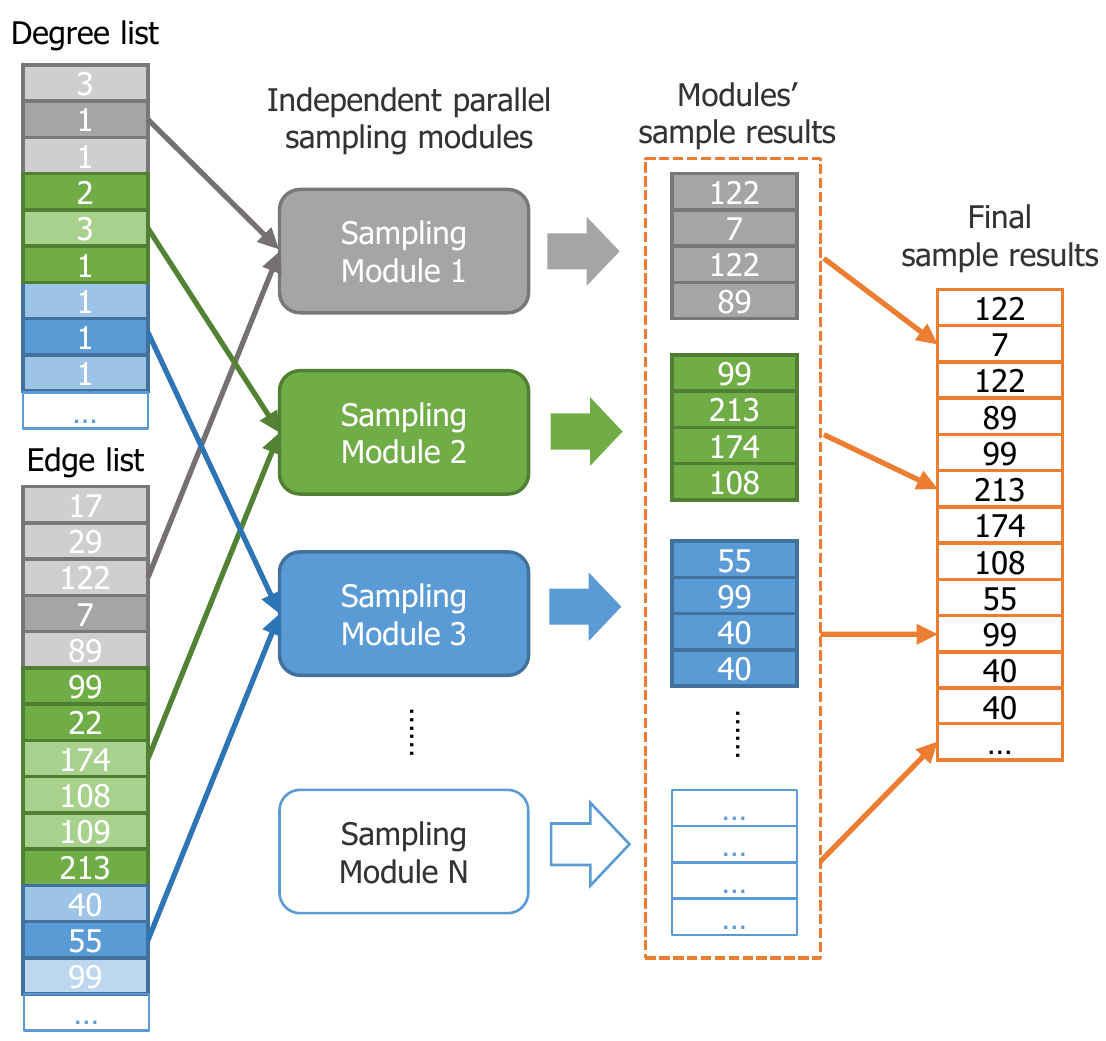}}
    \caption{Schematic diagram of parallel sampling process. Degree list and edge list are split into several segments, and for each module it only samples from the corresponding segment. We can concatenate each module's sample result into final sample result.}
    \label{parallel}
\end{figure}

In fact, for a single sampling module, it only consumes small amount of on-chip resources. Therefore, we can implement multiple sampling modules on our device in order to utilize all on-chip resources. Each sampling module can carry out sampling process independently, and their data input and output are in a parallel state. For dataset that needs to be sampled, we can divide the dataset beforehand, which means that we will divide its degree list and edge list into several segments, and each module independently sample from the corresponding segment, as shown in Fig.\ref{parallel}. Each module's sample result is output independently and can be directly concatenated into a complete sample result. Based on this idea, the sampling process can be further accelerated.


Here we give an example of dynamic power consumption data of the parallel sampling modules for small-scale datasets (use Cora as an example), as shown in the table \ref{power_table}. In our platform, 16 parallel modules can fully use the on-chip resources. It is important to note that the power consumed by the BRAM is related to the size of the dataset and the size of the segment, and the proportion of BRAM power consumption will increase if the size of degree list and edge list becomes larger.



\begin{table}[]
\caption{Dynamic power consumption of parallel sampling modules (Cora)}
\centering
\begin{tabular}{|c|c|c|}
\hline
\textbf{Dynamic Power Consumption Type} & \textbf{Power (W)} & \textbf{Percentage} \\ \hline
clocks                                  & 0.208              & 5\%                 \\ \hline
signals                                 & 0.736              & 16\%                \\ \hline
logic                                   & 0.48               & 11\%                \\ \hline
BRAM                                    & 0.16               & 4\%                 \\ \hline
others                                  & 2.912              & 64\%                \\ \hline
\end{tabular}
\label{power_table}
\end{table}

\section{Conclusion}
In this work, we proposed a novel, dedicated to hardware acceleration neighbor sampler ---- CONCAT Sampler. It can easy to be accelerated on hardware platform, and ensure the model's test accuracy at the same time. We also designed and hardware architecture to accelerate sampling process of CONCAT sampler. For CONCAT Sampler, its algorithmic features allow it to only sample 1-neighborhood to achieve n-neighborhood sampling, making its hardware acceleration architecture very simple and easy to implement.

In our experiment, we implemented our architecture on FPGA, and shortened sampling time by a factor of 300 to 1000. Meanwhile, we ensured that the final test accuracy of the model is consistent with the existing baseline, even improved a bit in some cases. For larger datasets, the speedup of our sampler is more pronounced. In fact, larger datasets often take more time in sampling process. Therefore, our proposed sampler is a good solution to slow sampling phenomenon, especially on large datasets.


\end{document}